\documentclass[11pt,a4paper]{article}
\usepackage[hyperref]{acl2021}
\usepackage{times}
\usepackage{latexsym}

\usepackage{microtype}

\aclfinalcopy 

\usepackage{xspace}
\usepackage{subfig}
\usepackage{booktabs}
\usepackage{makecell}
\usepackage{amsmath}
\usepackage{amssymb}
\usepackage{multirow}
\usepackage{xcolor}
\usepackage[switch]{lineno} 
\usepackage{graphicx} 
\usepackage{fixltx2e}

\newcolumntype{L}[1]{>{\raggedright\let\newline\\\arraybackslash\hspace{0pt}}m{#1}}
\newcolumntype{C}[1]{>{\centering\let\newline\\\arraybackslash\hspace{0pt}}m{#1}}
\newcolumntype{R}[1]{>{\raggedleft\let\newline\\\arraybackslash\hspace{0pt}}m{#1}}

\newcommand{\bertb}{BERT\textsubscript{BASE}\xspace}
\newcommand{\bertl}{BERT\textsubscript{LARGE}\xspace}

\newcommand{\data}{\textsc{ForecastQA}\xspace}
\newcommand{\ability}{\textit{forecasting ability}\xspace}
\newcommand{\para}[1]{\smallskip\noindent\textbf{#1}}

\title{\data: A Question Answering Challenge \\for Event Forecasting with Temporal Text Data}
\author{
    Woojeong Jin$^1$ \quad Rahul Khanna$^1$  \quad Suji Kim$^1$ \quad Dong-Ho Lee$^1$ \\ \textbf{Fred Morstatter}$^2$ \quad \textbf{Aram Galstyan}$^2$ \quad \textbf{Xiang Ren}$^1\,^2$ \\
    $^1$Department of Computer Science, University of Southern California\\
    $^2$Information Sciences Institute, University of Southern California\\
    \{woojeong.jin, rahulkha, sujikim, donghole, xiangren\}@usc.edu, \{fredmors, galstyan\}@isi.edu
}

\date{}

\begin{document}
\maketitle
\begin{abstract}

Event forecasting is a challenging, yet important task, as humans seek to constantly plan for the future. Existing automated forecasting studies rely mostly on \textit{structured data}, such as time-series or event-based knowledge graphs, to help predict future events.
In this work, we aim to formulate a task, construct a dataset, and provide benchmarks for developing methods for event forecasting with large volumes of \textit{unstructured} text data.
To simulate the forecasting scenario on temporal news documents, we formulate the problem as a restricted-domain, multiple-choice, question-answering (QA) task. 
Unlike existing QA tasks, our task limits accessible information, and thus a model has to make a forecasting judgement.
To showcase the usefulness of this task formulation, we introduce \data, a question-answering dataset consisting of 10,392 event forecasting questions, which have been collected and verified via crowdsourcing efforts.
We present our experiments on \data using BERT-based models and find that our best model achieves 60.1\% accuracy on the dataset, which still lags behind human performance by about 19\%.
We hope \data will support future research efforts in bridging this gap.\footnote{\small \url{https://inklab.usc.edu/ForecastQA/}}


\end{abstract}

\section{Introduction}
\vspace{-0.1cm}

\begin{figure}[tb!]
    \centering
    \includegraphics[width=0.96\linewidth]{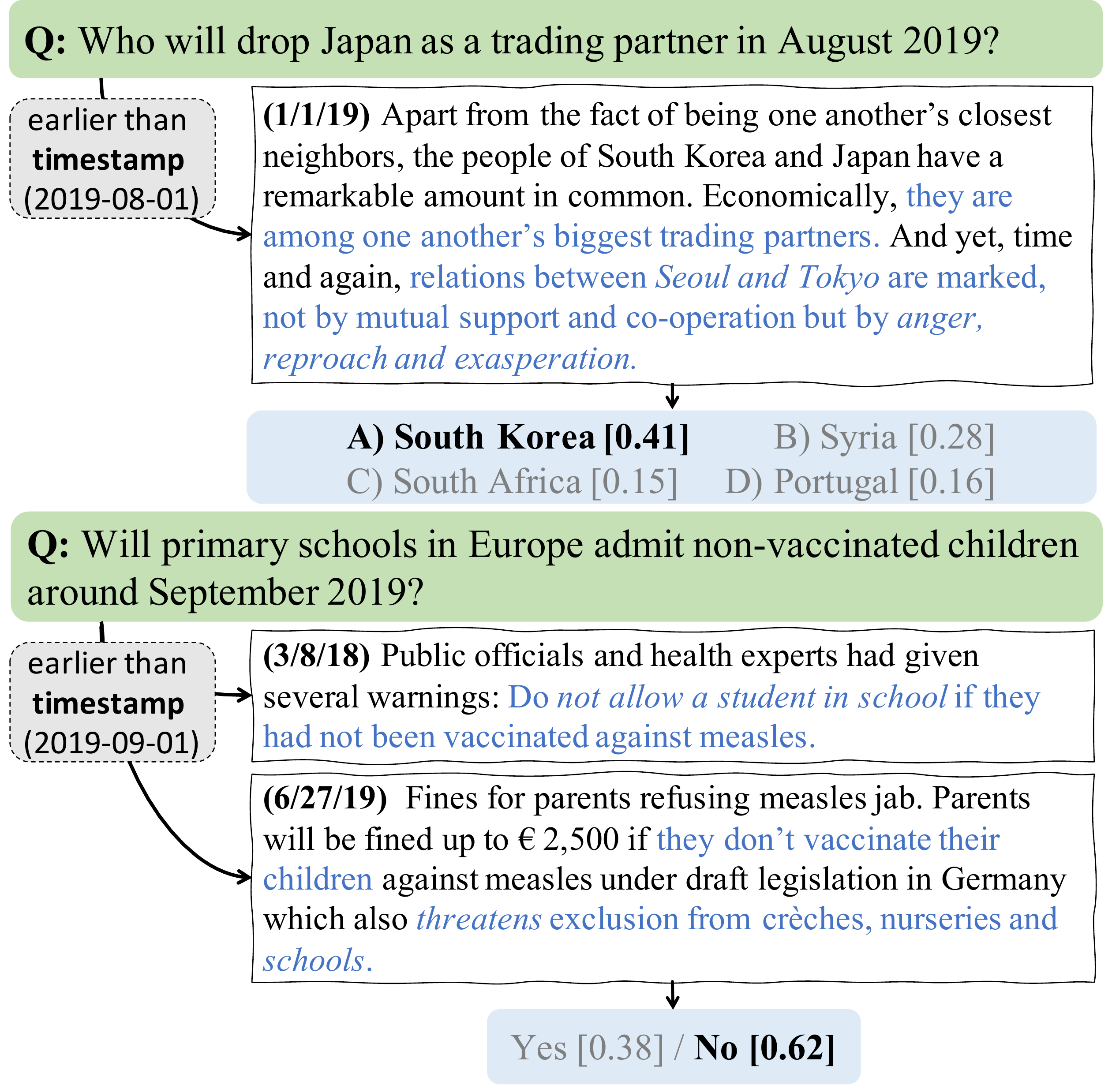}
    \caption{\textbf{Examples from the \data dataset.} Models only have access to articles published prior to the \textit{timestamp} associated with each question. 
    Models assign probabilities to each answer choice; bold denotes the correct answer for each question.
}
    \label{fig:example}
\end{figure}

Forecasting globally significant events, such as outcomes of policy decisions, civil unrest, or the economic ramifications of global pandemics, is a consequential but arduous problem. In recent years there have been significant advances in applying machine learning (\textit{e.g.}, time-series prediction methods) to generate forecasts for various types of events including conflict zones~\cite{schutte_2017}, duration of insurgency~\cite{Pilster2014}, civil unrest~\cite{embers} and terrorist events~\cite{raghavan2013}. 

Current automated forecasting methods perform well on problems for which there are sufficient \textit{structured} data (\textit{e.g.}, knowledge graphs), but are not well suited for events for which such data may not exist. Humans, though, can often accurately forecast outcomes by leveraging their judgement, domain knowledge, and prior experience~\cite{tetlock2016superforecasting}, along with the vast amounts of \textit{unstructured} text data available to us (\textit{e.g.}, news articles). We are able to identify and retrieve salient facts from the near-endless pool of unstructured information, synthesize those facts into coherent beliefs, and generate probabilistic forecasts. 
Unfortunately, the process does not scale well in terms of the amount of information that must be processed and the number of events one has to forecast.

Here we address the above problem by formalizing a forecasting task, creating a dataset, and providing benchmarks to develop methods for the task.
Specifically, we formulate the forecasting problem as a multiple-choice Question Answering (QA) task, where the input is a news corpus, questions, choices and timestamps associated with each question, and the output is one of the given choices per question. Our approach is rooted in the observation that both forecasting and QA follow a similar process: digesting massive amounts of textual data, identifying supporting pieces of evidence from text, and chaining different pieces to generate answers/forecasts. 

Forecast Question Answering (\data) introduces a novel \textit{timestamp constraint} per question that prohibits the model from accessing new articles published after the \textit{timestamp}. By doing so, \data simulates a forecasting scenario; each question's timestamp is chosen to ensure that the question is about the outcome of a future event. 

To illustrate this, consider the question, ``\textit{Will primary schools in Europe admit non-vaccinated children around September 2019?}" in Figure~\ref{fig:example}, and the fact that models only have access to articles before ``2019-09-01."
With the addition of this timestamp constraint, our query becomes a question about a future event in ``September, 2019" based on articles from the ``past"; the model is now being tested for its \ability\footnote{\scriptsize The ability to predict the outcome of future events based on unstructured text describing past events, without access to an extracted sequence of historical event triples, nor provided a fixed set of possible relations between events; as is the case with human forecasters.}. 
To answer the question, the model must find pertinent events from ``past" information, resolve the temporal and causal relations between them, and finally make a \textit{forecasting judgement} based on its interpretation of past information to answer the question.
Our task differs from that of other works that require an understanding of temporal relationships~\cite{ning2020torque} and temporal commonsense reasoning~\cite{Zhou2019GoingOA}, as our task forces a model to make a forecasting judgement.

In support of the proposed \data formulation, we construct a dataset of 10,392 yes-no and multiple-choice questions.
This data is collected via crowdsourcing based on news articles, where workers are shown articles and asked to come up with yes-no and multiple-choice questions. We also crowdsourced appropriate timestamps for each question.
Finally, we design a method based on pre-trained language models to deal with retrieved articles for our task.
In our experiments, the methods using retrieved articles slightly outperform closed-book models, suggesting that our task is still challenging in that finding relevant information for forecasting and making a judgement are not straightforward.
Our best attempt achieves 60.1\% accuracy on our dataset, a significant performance gap from human performance by 19.3\%.

\begin{table}[tb!]
\centering
\resizebox{1\linewidth}{!}{
\begin{tabular}{p{10cm}}
\toprule
\textbf{Q:} Who will drop Japan as a trading partner in August 2019?  \\
\textbf{Choices:} South Korea (\textit{\textbf{answer}}), South Africa, Syria, Portugal. \\
\cmidrule[1pt]{1-1}
\textbf{Article:}  \textit{Why Japan and South Korea just can’t get along.} (1/1/19) \newline Apart from the fact of being one another’s closest neighbours, the people of South Korea and Japan have a remarkable amount in common. Economically, they are among one another’s biggest trading partners. And yet, time and again, relations between Seoul and Tokyo are marked, not by mutual support and co-operation but by anger, reproach and exasperation. \\
\cmidrule[0.5pt]{1-1}
\textbf{Reasoning Process:} Seoul is in South Korea, Tokyo is in Japan (\textbf{commonsense - world knowledge}). Seoul and Tokyo are big trading partners  (\textbf{language understanding - lexical variations}). The relations between Seoul and Tokyo are marked by anger, reproach and exasperation and these relations might cause trading relations to cease (\textbf{forecasting skills - causal relation} - \textit{we can infer the answer from this part}).
 \\\bottomrule
\end{tabular}}
\caption{\textbf{Chain of reasoning.} The question requires the reasoning process to answer.
}
\label{table:detailed_example}
\end{table}

\section{Related Work}


\noindent
\textbf{Event Forecasting.}
There are several types of approaches exist to do event forecasting. One approach could learn from highly structured event-coded data such as ICEWS~\cite{boschee2015icews} and GDELT~\cite{leetaru2013gdelt}.
When these datasets are used for forecasting, they are often represented as a time series~\cite{Morstatter2019SAGEAH,ramakrishnan2014beating}, in which each data point is associated with a timestamp. 
Another approach is script-learning, in which a model is provided with a chain of events and a subsequent event and is asked to predict the relation between the chain and the ``future" event ~\cite{hu2017happens,li2018constructing,lv2019sam}. 
They require to convert text data into event triples and translate the questions and answer choices into their format, which limits the expressiveness of natural text.
However, unlike these datasets and approaches, \data does not provide any structured data to a model. The model must learn how to extract, keep track of, and link pertinent events from unstructured text to solve forecasting questions.


\para{QA and Temporal Reasoning on Text.}
There are several approaches for QA using unstructured text. Extractive QA approaches rely on finding answer spans from the text that best answer a question~\cite{Rajpurkar2016SQuAD10,Rajpurkar2018KnowWY,Yang2018HotpotQAAD,kwiatkowski2019natural,huang2019cosmos}. 
Multiple-Choice QA requires a model to pick the best answer from a set~\cite{Talmor2018CommonsenseQAAQ,Sap2019SocialIC,Zhou2019GoingOA}, and generative QA prompts the machine to produce its own answer~\cite{khashabi2020unifiedqa}. 
Our dataset is a type of multiple-choice QA, but it differentiates itself from other  QA datasets (all formats) in that the required answer does not exist in the provided text, nor is sufficient evidence provided to be able to answer a question with 100\% certainty; a forecast is required.
We could convert our questions into alternative query formats such as a text-to-text format, but instead we stick to multiple-choice questions as humans often weigh the benefits of multiple choices when making a forecasting judgement.

QA datasets often exist to test certain types of reasoning. One pertinent example of a reasoning type that QA tasks test is the understanding of temporal and casual relations~\cite{Jia2018TempQuestionsAB,Jia2018TEQUILATQ,Sun2018ReadingCW,ning2020torque}. However, \data requires more than just extraction and understanding of relations; a model must be able to extract and understand the relations present in the text with the goal of making a forecasting judgement about an event whose outcome is \textit{not} found in the text.
Another type of reasoning tested in QA tasks is commonsense reasoning~\cite{Talmor2018CommonsenseQAAQ} and even temporal commonsense reasoning~\cite{Zhou2019GoingOA}. While questions in \data often require commonsense to correctly answer, not all do; event outcomes do not always follow common sense. 
Furthermore, our questions test forecasting abilities, which often includes various types of reasoning in addition to commonsense.


\section{The \data Task}
\label{sec:data}

\begin{figure}[tb!]
    \centering
    \includegraphics[width=1\linewidth]{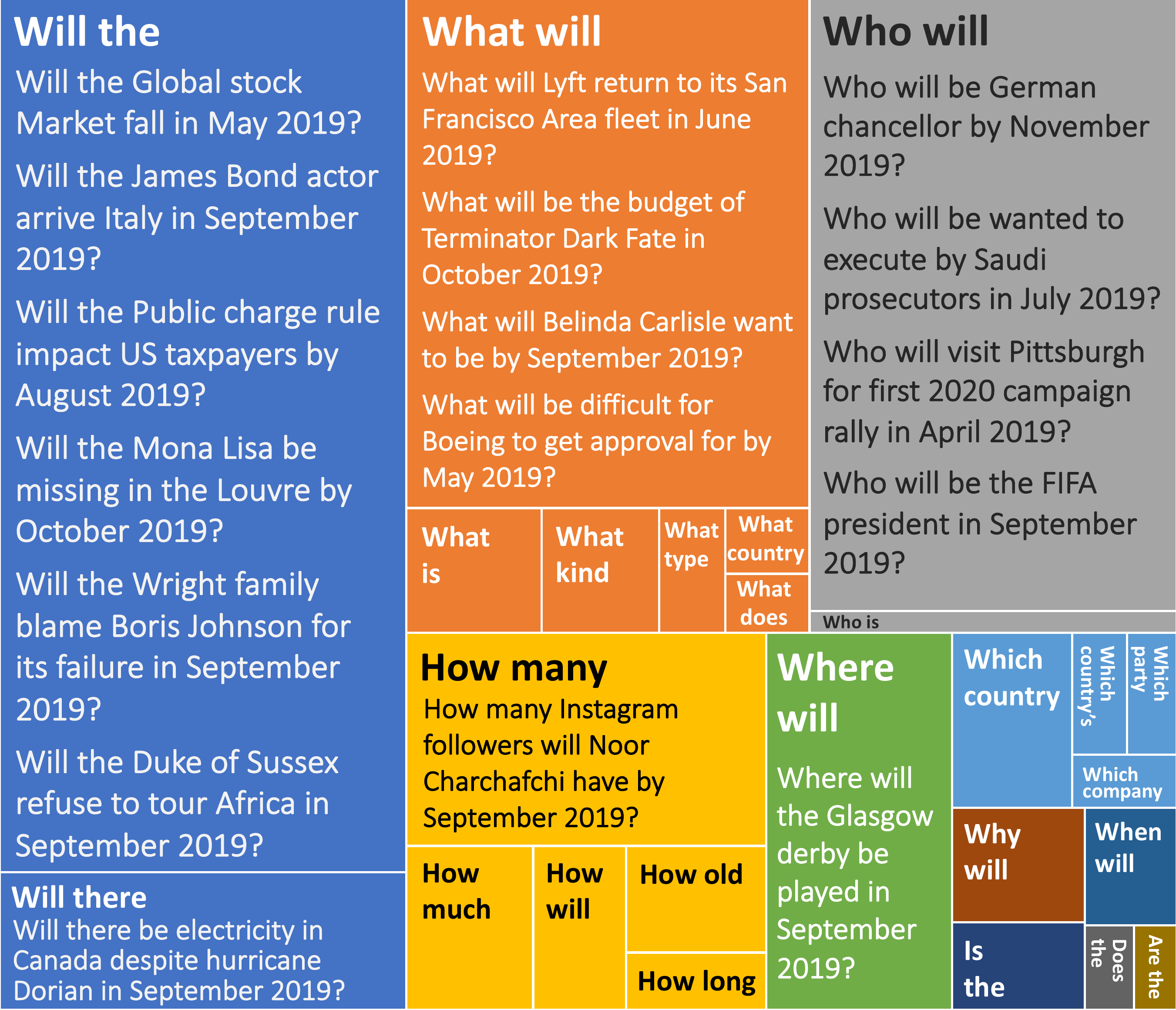}
    \caption{ \textbf{A treemap visualization of first two words in \data questions.} Box area is proportional to number of occurrences. 
}
    \label{fig:distquestions}
\end{figure}

\data is a question answering task whose goal is to test a machine's \ability. 
We consider \textit{forecasting} as the process of anticipating the outcome of future events based on past and present data~\cite{tetlock2016superforecasting}.
We focus on forecasting \textit{outcomes of news-based events} coming from topics such as politics, sports, economics, etc. 
Training a machine to make forecasting decisions is inherently difficult, as the ground-truth label of event outcome (\textit{e.g.}, whether an event will occur) --- so often required for model training --- is only obtainable ``in the future". 
To make progress in our goal, 
we devise a way to \textit{simulate the forecasting scenario} by introducing a novel \textit{time constraint}, allowing us to validate the machine predictions by obtaining desired ground-truth labels.

There is also the difficulty of ensuring the quality of question generation via crowdsourcing (necessary when building a dataset of scale), due to possible human errors in question formation~\cite{Tetlock2017BringingPJ}.
We have taken steps to ensure our questions cannot be answered \textit{with certainty} using ``past" data given the time constraint or commonsense knowledge, but the questions are \textit{tractable} to answer with an educated guess
(see Sec.~\ref{subsec:data_construction}).\footnote{\small This is in contrast to open-domain QA (machine reading comprehension)~\cite{kwiatkowski2019natural} where answers can always be found in some given passages.} 

\begin{table}[tb!]
\centering
\resizebox{0.87\columnwidth}{!}{
\begin{tabular}{l|ccc|c}
\toprule
\textbf{Statistic}    &\textbf{Train}  & \textbf{Dev} &\textbf{Test}   &\textbf{All}\\
\cmidrule[1pt]{1-5}
Questions&8,210   &1,090  &1,092  &10,392\\
\cmidrule{1-5}
Yes-no questions&4,737   &582 & 584   &5,903\\
Multi-choice questions&3,473  &508 & 508   &4,489\\
\bottomrule
\end{tabular}}
\caption{\small \textbf{Size of the \data dataset.}}
\label{data:stats}
\end{table}

\para{Task Definition.}
\label{subsec:taskdef}
Formally, the input of the \data task is a forecasting question $Q$ with a corresponding ending timestamp $t_Q$––the last possible date where $Q$ remains a forecasting question. In addition, we have a set of possible choices, $\mathcal{C}$, and a corpus of news articles, $\mathcal{A}$; the output is a choice $C \in \mathcal{C}$.
Our task has a novel \textit{constraint} that any retrieved article $A \in \mathcal{A}$ must satisfy $t_A < t_Q$. 
In other words, models have access only to articles that are published before $t_Q$. 
We have ensured that the information required to solve the question \textit{deterministically} comes out in an article, \textit{gold article}, published after $t_Q$, i.e., $t_\text{gold\_article} \geq t_Q$. Another way to think of our setup is that we are asking $Q$ on the day before $t_Q$, knowing that the information required to solve $Q$ is not available yet. 
This formulation makes our task both a constrained open-domain QA and a forecasting problem––distinct from existing QA tasks.

\para{Challenges in \data.}
Due to the constrained open-domain setting and forecasting properties, testing a model's \ability encompasses the following challenges:
information retrieval (IR) on limited sources, understanding of temporal and causal relations between events, and finally a forecasting judgement.
Our time constraint limits the accessible articles and also creates more challenges than in standard open-domain QA; effective IR methods are necessary to anticipate what knowledge will be useful for predictions from past information sources.
Once useful articles have been retrieved, models should understand these articles and reason over pertinent facts from them. 
Finally, these models use the gleaned knowledge to infer the outcome of a future event.
Unlike in other reading comprehension tasks, models cannot rely on the existence of an answer within the text, but must make an educated guess as to what will happen in the future. 
While our task does encompass reasoning abilities tested in other datasets, no other tasks investigate these reasoning abilities in the context of predicting future events.
More analysis on reasoning types can be found in Sec.~\ref{subsec:reasontype}.

\section{Dataset Construction and Analysis}
In this section, we describe how we construct our \data dataset and analyze it.

\begin{figure}[tb!]
    \centering
    \includegraphics[width=1\linewidth]{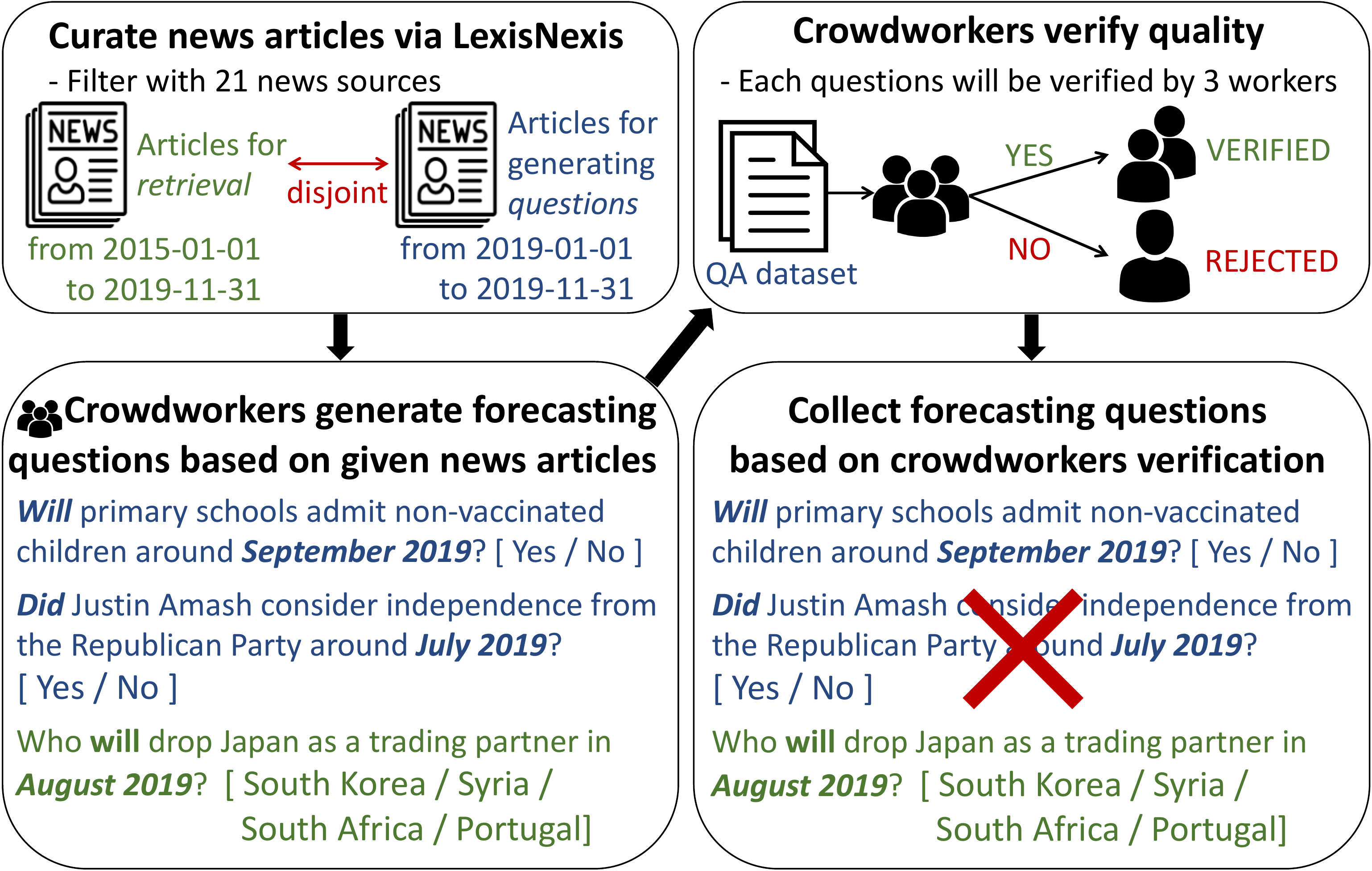}
    \caption{ \textbf{\data generation process.} The input of \data creation is a news article corpus and the output is yes-no/multiple-choice questions.
}
    \label{fig:generation}
\end{figure}

\subsection{Construction Details}
\label{subsec:data_construction}
The data collection is broken down into three sections: (1) gathering a news corpus, (2) generating question-answer-timestamp triples with distractor choices, and (3) verifying the triples' quality. 
The data generation process is summarized in Fig.~\ref{fig:generation}.

\para{News Corpus Collection.}
\label{data:news}
We started by gathering English news articles from LexisNexis\footnote{\scriptsize \url{https://risk.lexisnexis.com}}. 
We then curated a list of 21 trustful news sources and filtered articles based on their publishers; we also filtered out non-English articles. Finally, we selected the five-year period of 2015-2019 and filtered out articles outside this period, leaving us with 509,776 articles. This corpus is also used for retrieval in our task setting (\textit{i.e.}, constrained open-domain).

\para{Q-Answer-timestamp Triple Creation.\footnote{\scriptsize Due to the limited space, for more details of our triple creation guidelines for human annotators, verification steps, and screenshots of our data collection/verification AMT interfaces, please refer to Sec.~\ref{append:data} of the appendix.}}
Once we assembled the news corpus, we built (question, answer, timestamp) triples to accompany the new corpus as inputs for our task. To generate the needed triples we looked to crowdsourcing via Amazon Mechanical Turk. Our generation task consists of the following steps:
(1) we selected a random news article from 2019 from the collected news corpus (these news articles are \textit{gold articles} and will be hidden for experiments);
(2) workers created questions, which \textit{if posed before the respective article’s publication date} would be seen as a forecasting question;
(3) they indicated the answer, along with supporting evidence that the question consisted of (to ensure the correctness of the true answer);
(4) they were asked to make multiple-choice distractors with their own knowledge and/or access to search engines; and
(5) we ensured that a temporal phrase is present in the questions, for example: ``\textit{After May of 2020...}", ``\textit{... in June of 2021?}" to provide a temporal context (constraint) for each question, yielding more precise and well-defined forecasting questions. 
Completion of this task results in the desired triple of: a forecasting question, an answer to the question (with distractor choices), and a timestamp as our temporal constraint. The timestamp is set as the first day of the month in which the gold article was published.

\begin{figure*}[tb!]
    \centering
    \includegraphics[width=0.96\linewidth]{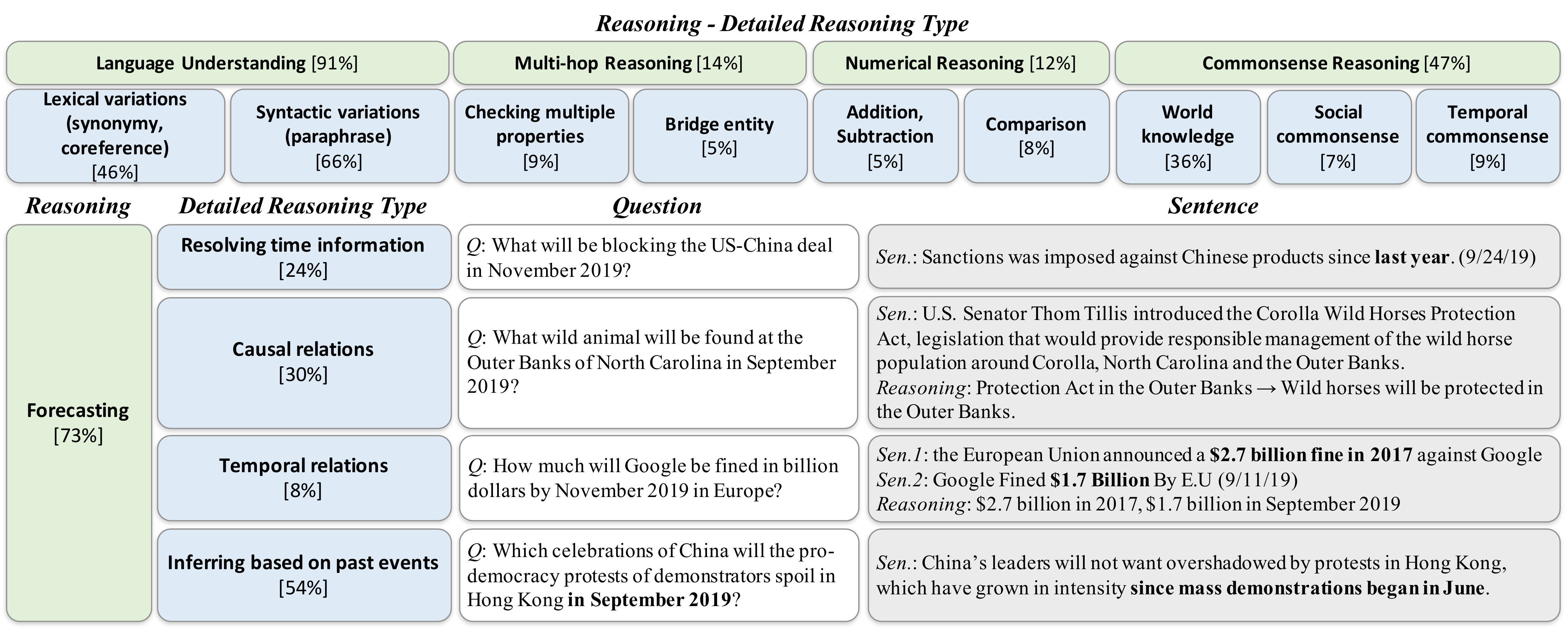}
    \caption{\small \textbf{Reasoning skills (types) and their frequency (in \%) in the sampled data.} As each question can be labeled with multiple types, the total frequency does not sum to 100\%.
    On average, 3 reasoning skills are required for each question. 
    Examples of other reasoning types can be found in Fig.~\ref{fig:reasoning_example2} in the appendix. 
    }
    \label{fig:reasoning_example}
\end{figure*}


To diversify questions in the dataset, we created two kinds of questions: binary yes-no questions and multiple-choice questions with \textit{four choices}.
Multiple-choice questions start with one of the six Ws (\textit{i.e.}, who, what, when, where, why, and how) and are more challenging as they require determining the correctness of each choice.

\para{Question Quality Verification.}
We performed a separate crowdsourcing data verification to test and enforce the following criteria: 
(1) is answering the question a \textit{tractable} problem given (relevant) ``past" articles?,
and (2) is the question \textit{deterministically} answerable given any article adhering to the question’s temporal constraint? ---
If a question is too difficult, \textit{i.e.}, an educated guess to the answer (when given relevant, constraint-adhering articles) is not possible, then we filter the question out. 
On the other hand, if the questions are answerable \textit{with certainty} using “past” articles, or commonsense/world knowledge, then they are \textit{not} considered to be forecasting questions. The desired response (majority vote from 3 annotators) is a ``\textit{yes}" for criterion (1) and ``\textit{no}" for (2), as that would show that the tuple of question and time constraint simulates the desired forecasting scenario. 
With the above method, we filtered out 31\% of the questions collected in the triple creation step and were left with 5,704 yes-no questions and 4,513 multi-choice questions. More details about the verification step are included in Sec.~\ref{append:data} of the appendix.

\subsection{Dataset Analysis}
\label{sec:reason}
To better understand the properties of the questions in \data, we examine: 1) a few data statistics 2) types of questions asked, and 3) the types of reasoning required to answer our questions.

\para{Summary Statistics.}
\data dataset is composed of 10,392 questions, divided into a 80/10/10 split of train, dev, and test data. Our 10k questions are roughly evenly split between multiple-choice and yes-no binary questions (Table~\ref{data:stats}). Over 17K distinct words were used to construct our questions and we have 218 unique time constraints associated with them; time constraints range from 2019-01-11 to 2019-11-12.
We include additional statistics in Sec.~\ref{append:stat} of appendix.

\para{Types of Questions.}
To understand the types of questions in \data, we examined the popular beginnings of sentences and created a tree-map plot (see Fig.~\ref{fig:distquestions}).
As shown, nearly half the questions start with the word \textit{will} (44\%), a result of over half of the questions being yes-no questions.

\para{Reasoning Types.}
\label{subsec:reasontype}
To examine types of reasoning required to answer our questions we sampled 100 questions and manually annotated them with reasoning types. Due to the forecasting nature of our dataset, we are particularly interested in questions containing the forecasting ability and thus spend more time looking into these questions. Our condensed results can be found in Figure~\ref{fig:reasoning_example}, and more results from our cataloguing effort can be found in  Sec.~\ref{appendix:reasoningtypes} of the appendix. Note that most questions contain more than one reasoning type.

\section{Methods}
\label{sec:models}
To evaluate the forecasting capabilities of recent \textit{multi-choice}/\textit{binary} QA model architectures on \data, we provide a comprehensive benchmarking analysis in this work.   
We run the experiments in two settings: (1) \textit{closed-book} and (2) \textit{constrained open-domain} setup.
In the \textit{closed-book} scenario only $Q$ (question) and $C$ (answer choices) are provided to the model $(Q,C)$, while $\overline{A}$ (news articles) is provided for setting (2), $(Q,C,\overline{A})$\footnote{\scriptsize $t_Q$ is always applied to $\overline{A}$, we left it out of the notation for simplicity.}. 
We run these settings to understand the difficulty of both the closed-book and open-domain challenges presented by the questions in \data. 

\begin{figure}[tb!]
    \centering
    \includegraphics[width=0.98\linewidth]{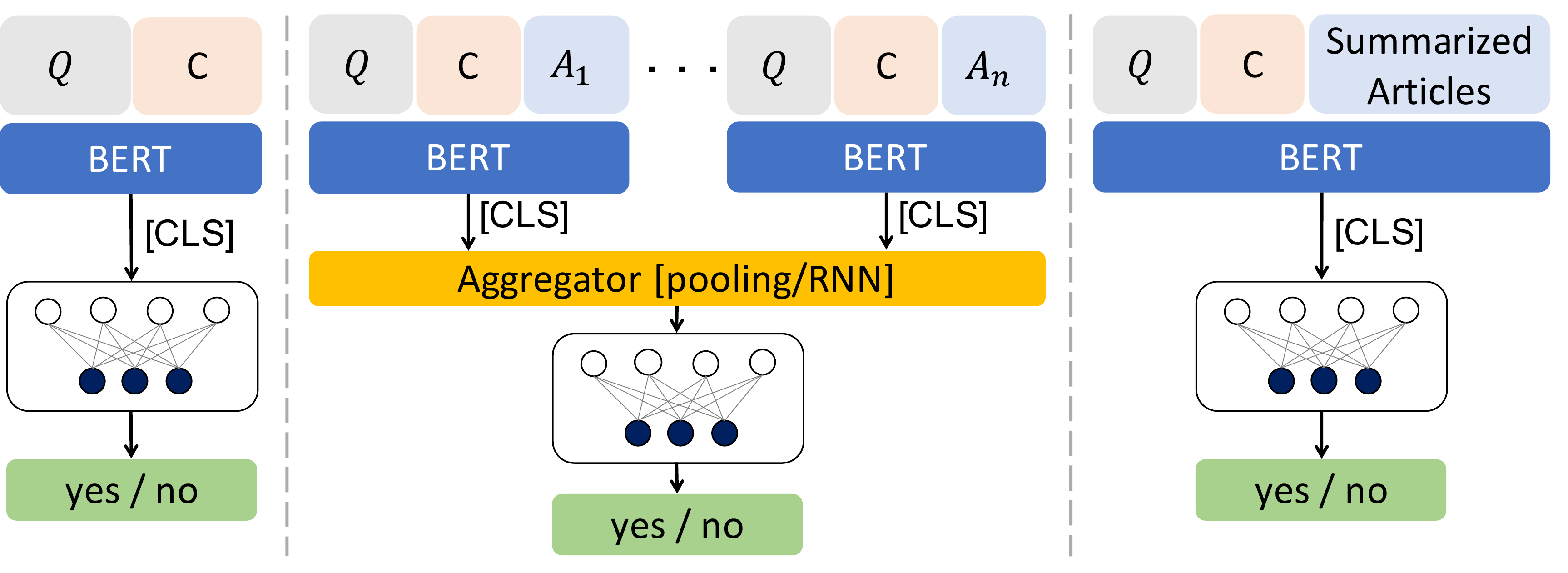}
    \caption{\small \textbf{Our baseline model architectures.} The CLS token is either fed into an MLP for classification or to the aggregator, which collects the information from each article before classifying. 
}
    \label{fig:structure}
\end{figure}

For both settings, we explore several baseline models, but all follows a general architecture of a text encoder $f$ and an optional context aggregation module $g$ to aggregate information from a set of retrieved articles. Fig.~\ref{fig:structure} shows the architectures used. 
We model both yes-no and multiple-choice questions as a \textit{binary} classification task; a model's prediction is the class with the largest probability.
Below we introduce the details of our baselines.

\para{Text Encoder.}
We use pre-trained language model, BERT~\cite{Devlin2019BERTPO}, as a text encoder ($f$ from above)\footnote{{\scriptsize We did not include more recent pre-trained language models (\textit{e.g.}, RoBERTa~\cite{Liu2019RoBERTaAR}, ALBERT~\cite{lan2019albert}, T5~\cite{raffel2020exploring}) or pre-trained QA models like UnifiedQA~\cite{khashabi2020unifiedqa}, as these models are trained using text data published \textit{after} the earliest timestamp in our dataset (2019-01-01), meaning information leakage could occur (and violates the forecasting setup). We tested more LMs in Sec.~\ref{append:stronglms} of appendix.}}.
$f$ is designed to deal with $(Q,C)$ and $(Q,C,\overline{A})$ inputs, where $\overline{A}$ is a set of time-stamped articles that are retrieved from $\mathcal{A}$ to answer $Q$. Each input of $f$ is transformed into 
$[\texttt{[CLS]} Q \allowbreak \texttt{[SEP]} C \texttt{[SEP]} A_i]$ (for each $A_i \in \overline{A}$, $C \in \mathcal{C}$), or $[\texttt{[CLS]} Q \allowbreak \texttt{[SEP]} C]$ (for each $C \in \mathcal{C}$) if articles are not supplied. The \texttt{[CLS]} token is the same as the one commonly used for fine-tuning PTLMs for a classification task, and \texttt{[SEP]} is the special separator token.
The embedding of \texttt{[CLS]} is then used for predictions with an MLP layer (the leftmost model architecture in Fig.~\ref{fig:structure}), or as input into a context aggregation module (the middle architecture in Fig.~\ref{fig:structure}) subsequently introduced.

\para{Context Aggregation (AGG).}
Two architectures are used when aggregating information from multiple, time-stamped articles $\overline{A}$ retrieved for a question. 
(1) \textit{Temporal Aggregation:} This aggregator utilizes temporal ordering of the retrieved articles. 
Articles are sorted by their timestamps and their \texttt{[CLS]} token representation from $f$ are aggregated by a Gated Recurrent Unit (GRU)~\cite{Cho2014LearningPR} with a MLP head to make final predictions. 
(2) \textit{Set Aggregation:} Alternatively, we ignore the temporal ordering of articles and use a \textit{maxpooling operation} on the \texttt{[CLS]} token representations of each article. This pooled representation is passed to an MLP layer to make a prediction. 
Comparison between these aggregations helps understand the effect of modeling temporal order of evidence. These two aggregation modules are denoted by ``AGG (GRU)" and ``AGG (Maxpool)," respectively. 

\para{Multi-document Summarization (MDS).}
Rather than conducting context aggregation of the retrieved articles, we consider an MMR summarizer~\cite{carbonell1998use} which performs extractive, multi-document summarization of text to generate a summary $A_{summ}$ (rightmost architecture in Fig.~\ref{fig:structure}).
The summary article $A_{summ}$ is treated as if it is an $A_i \in \overline{A}$ and fed into a text encoder along with $Q$ and $C$ which then produce the \texttt{[CLS]} embedding for making a prediction. We name this method ``MDS."

\para{Integrated Approach.}
To take the best of both worlds in $(Q,C)$ and $(Q,C,\overline{A})$ settings,
we integrate two architectures (the leftmost and middle ones in Fig.~\ref{fig:structure}).
We concatenate the last two hidden representations of each architecture before passing the concatenated representation through a shared MLP layer. We use \bertl as $f$ in both architectures, AGG (GRU) for $g$ and call this model ``\bertl++ (integrated)" in Table~\ref{result:open}.

\para{Other Baselines.}
We also consider other baselines: ESIM~\cite{Chen2017EnhancedLF}, \textsc{BiDAF++}~\cite{Clark2018SimpleAE}, prepending extracted open event triples~\cite{liu2019open} to BERT input, and a script learning approach, SAM-Net~\cite{lv2019sam}. We modify the approaches to fit into our setup. Detailed descriptions of each baseline method are included in Sec.~\ref{exp:baselines} of appendix. 

\section{Experiments}
\subsection{Experimental Setup}
\label{sec:expset}
We adopt two types of settings: the closed-book setting $(Q,C)$ and the constrained open-domain setting $(Q,C,\overline{A})$. In the constrained open-domain setting, we use BM25~\cite{robertson1995okapi,Qi2019AnsweringCO} as our IR method\footnote{\small Details of IR methods are described in appendix Sec.~\ref{exp:ir}.} to obtain $\overline{A}$, 10 retrieved articles. We also explore other IR methods in the later section. Note that we retrieve articles that do not violate the time constraints. We feed the question $Q$ as a query and limit our access to articles in $\mathcal{A}$ by $t_Q$.
Additionally, we validate the answerability of our questions by providing gold articles 
instead of retrieved articles (Sec.~\ref{subsec:result_analysis}). 

\para{Evaluation Metrics.}
Because forecasting is uncertain, a system's prediction probabilities indicate its confidence  answering the question.
In addition to accuracy, we consider Brier score~\cite{brier1950verification}, which measures the mean squared \textit{error} of probabilities assigned to sets of answer choices (outcomes). Formally, $\text{Brier}=\frac{1}{N} \sum_{i=1}^N \sum_{c=1}^C (p_{ic} - y_{ic})^2$,
where $p_{ic}$ is the probability of prediction; $y_{ic}$ is a label indicator for class $c$ of the instance (1 or 0), $N$ is the number of prediction instances, and $C$ is the number of classes (2 or 4). The highest Brier score is 0 (probability 1 for the correct class, probability 0 else), while the worst possible Brier score is 2 (probability 1 for the wrong class, probability 0 else).
A confident model gets low Brier scores.

\begin{table}[tb!]
\centering
\resizebox{1.0\linewidth}{!}{
\begin{tabular}{lcccccc}
\cmidrule[1pt](){1-7}
\multicolumn{1}{c}{\multirow{1}{*}{\hspace{0mm}\vspace{-7mm}\textbf{Methods / Metrics}}} & \multicolumn{3}{c}{\textbf{Accuracy (\%,~$\uparrow$)}} & \multicolumn{3}{c}{\textbf{Brier score ($\downarrow$)}}\\
\cmidrule(lr){2-4} \cmidrule(lr){5-7}
         &\textit{yes/no}&\textit{multi} &\textit{all} &\textit{yes/no} &\textit{multi} &\textit{all}     \\
\cmidrule[1pt]{1-7}
Random  &48.6    &25.3      &37.8      &0.684      &0.827      &0.750 \\
\midrule
ESIM-ELMo (closed-book) &63.3  &45.8      &54.5      &0.515      &0.897      &0.706 \\
\bertb (closed-book) &66.2 &41.5      &54.7      &0.511      &0.715      &0.606 \\ 
\bertl (closed-book) &67.3 &45.4      &57.6      &0.447      &0.653      &\textbf{0.543} \\
\midrule
\textsc{BiDAF}++~\cite{Clark2018SimpleAE} 
&51.7    &30.1     &40.9    &0.478      &0.898      &0.688 \\
\bertb, MDS  &63.1    &39.1      &52.0      &0.504      &0.716      &0.603 \\

\bertb, AGG (Maxpool)  &67.2      &39.1      &54.2      &0.453      &0.701      &0.568 \\
\bertb, AGG (GRU)     &67.6    &41.5      &55.4      &0.477      &0.705      &0.583 \\
SAM-Net~\cite{lv2019sam}   &64.5  &40.9      &53.5      &0.531      &0.719      &0.619 \\
\bertl, MDS   &67.4   &40.1   &54.7   &0.542   &0.738    &0.633     \\
\bertl, Event triples      &66.7  &45.0      &56.6      &0.589      &0.719      &0.649 \\
\bertl, AGG (Maxpool)   &68.8   &46.9   &58.6   &0.476   &\textbf{0.648}    &0.556 \\
\bertl, AGG (GRU)  &69.2   &47.5   &59.1   &0.483   &0.655    &0.563    \\
\bertl, AGG (Maxpool), DPR  &70.2  & 47.0   &59.4  &0.554   &0.728    &0.635    \\
\bertl, AGG (Maxpool), BT   &70.0    &48.0   &59.7   &\textbf{0.444}   &0.662   &0.545\\
\bertl++ (integrated) &\textbf{70.3}   &\textbf{48.4}   &\textbf{60.1}   &0.537   &0.650    &0.589     \\
\midrule
Human performance$^{(\alpha)}$ &74.6      &64.9      &71.2      &-      &-      &-\\
Human performance$^{(\beta)}$  & 81.3      & 77.4      & 79.4      &-      &-      &-\\
\cmidrule[1pt](){1-7}
\end{tabular}}
\caption{\small \textbf{Performance of baseline models on \data test set.} ``yes/no" refers to yes-no questions, and ``multi" to multi-choice questions. We test the closed-book setting, and the constrained open-domain setting, where the accessible articles are limited by $t_Q$, our time constraint.  
We use BM25 as the article retriever to select top-10 articles, if not particularly specified.
``BT" concatenates the binary encoding of date string to an article encoding before aggregation (see Sec.~\ref{subsec:result_analysis} ``Ablation on Timestamp Modeling"). 
Human performance is based on the top-10 retrieved articles $({\alpha})$, and Google Search with the question's time constraint $({\beta})$. 
}
\label{result:open}
\end{table}

\subsection{Human Performance}
To benchmark human performance, seven annotators (computer science graduate students) who were not involved in question generation were asked to answer 150 randomly sampled questions from the test set.
We consider two scenarios: 1) annotators are provided with retrieved articles, $\overline{A}$; and 2) annotators can access any article published \textit{before the timestamp} via Google Search. 
Moreover, as annotators live in the ``future" with respect to the timestamp of a question, they might already know the actual answer. To avoid the over-estimation of accuracy, we asked the annotators to not use their ``future" knowledge. If they felt this is not possible, we asked them to skip the question. On average, 28.3\% of questions are skipped.
Given this setup, humans achieve 71.2\% and 79.4\% accuracy respectively, for the two scenarios when taking a majority vote for each question; we also observed good inter-annotator agreement.
The two scenarios are referred as ``$(\alpha)$" and ``$(\beta)$" in Table~\ref{result:open}.

\subsection{Results and Performance Analysis}
\label{subsec:result_analysis}
\vspace{-0.1cm}


\noindent
\textbf{Results on the Constrained Open-domain Setting.}
Table~\ref{result:open} shows the results of baseline methods for comparison.
We compare pre-trained language models with different context aggregators and other baselines.
The integrated model, \bertl++ shows the best performance in terms of accuracy, while \bertl (closed-book) shows the best Brier score.
Unlike the accuracy metric, the Brier score penalizes over- and under- confident forecasts~\cite{mellers2014psychological} --- thus the best model under each metric can be different. The marginal differences in performance between the two settings suggest that access to information (text evidence) alone does not solve the forecasting problem. We hypothesize an inability to encode salient relations for forecasting purposes prevents the additional information from proving useful.
Among the aggregators in \bertb, the GRU aggregator outperforms other aggregators and summarizers. 
This suggests that utilizing articles' temporal order helps the reasoning. 
Overall, baselines fall behind human performance by over 10\% points given the same retrieved articles.

\begin{table}[tb!]
\centering
\resizebox{0.82\linewidth}{!}{
\begin{tabular}{lC{1.4cm}C{1.4cm}C{1.4cm}}
\cmidrule[1pt](){1-4}
\textbf{Methods} &GRU & Maxpool & MDS    \\
\cmidrule[1pt]{1-4}
\bertb, TF-IDF  &53.2   &53.9   &51.6   \\
\bertb, DPR &53.7   &\textbf{54.6}   &\textbf{54.3}   \\
\bertb, BM25    &\textbf{55.4}   &54.2   &52.0   \\
\midrule
\bertl, TF-IDF &56.5   &55.4   &\textbf{55.0}  \\
\bertl, DPR     &56.1   &\textbf{59.4}   &54.6   \\
\bertl, BM25    &\textbf{59.1}   &58.6   &54.7   \\
\cmidrule[1pt](){1-4}
\end{tabular}}
\caption{\small \textbf{Accuracy with different retrievers:} BM25, TF-IDF, and dense passage retrieval (DPR). We test the retrievers with different aggregators: GRU, Maxpool, and MDS.
}
\label{result:retriever}
\end{table}
			

\para{Study of Different IR Methods.}
We further test several retrieval methods: BM25~\cite{robertson1995okapi,Qi2019AnsweringCO}, TF-IDF~\cite{Chen2017ReadingWT}, and a pre-trained dense passage retriever (DPR)~\cite{karpukhin2020dense}.
As in Table~\ref{result:retriever}, \bertl with DPR retriever and the Maxpool aggregator shows the best performance than other combinations.
However, DPR does not achieve the best accuracy for all methods.
This implies that 1) stronger retrieval methods are required to identify useful evidence; 2) complex forecasting abilities may be a bottleneck of current systems.

\begin{table}[tb!]
\centering
\resizebox{1\linewidth}{!}{
\begin{tabular}{lccccc}
\cmidrule[1pt](){1-5}
\multicolumn{1}{c}{\multirow{1}{*}{\hspace{0mm}\vspace{-7mm}\textbf{Methods / Metrics}}} & \multicolumn{2}{c}{\textbf{GRU}} & \multicolumn{2}{c}{\textbf{Maxpool}}\\
\cmidrule(lr){2-3} \cmidrule(lr){4-5}
         &ACC ($\uparrow$)   &Brier ($\downarrow$)  &ACC ($\uparrow$)   &Brier ($\downarrow$)    \\
\cmidrule[1pt]{1-5}
w/o timestamps &\textbf{55.4}   &\textbf{0.583}   &54.2   &\textbf{0.568} \\
\midrule
Pre-pend timestamps  &54.2   &0.634   &54.8   &0.599 \\
Binary timestamp encoding       &51.1   &0.623   &\textbf{55.6}   &0.624 \\
Char-RNN timestamp encoding    &54.0   &0.640   &54.3   &0.620 \\
\cmidrule[1pt](){1-5}
\end{tabular}}
\caption{\small \textbf{Study on modeling article timestamps (publication dates) in the constrained open-domain setting.} We test several methods for temporal modeling and use \bertb with two different aggregators: GRU and Maxpool.
}
\label{result:date}
\end{table}

\para{Ablation on Timestamp Modeling.}
We conduct an ablation study on modeling time information (publication date) of the retrieved articles, as seen in Table~\ref{result:date}. 
We test: a) pre-pending date string as BERT input, b) using binary encodings of dates\footnote{\small \url{https://temporenc.org}} and concatenate with article encoding before aggregation, and c) using char-RNN~\cite{goyal2019embedding} for encoding date string before aggregation\footnote{Details are described in appendix Sec.~\ref{exp:time}}.
We find that using binary encodings of dates improves the accuracy for the maxpool aggregator.
However, the GRU aggregator's accuracy decreases when given date information. 
We conjecture that our modeling for the time information of each article is not strong enough to help forecasting. We leave more sophisticated modeling for future work.

\begin{table}[tb!]
\centering
\resizebox{0.97\linewidth}{!}{
\begin{tabular}{lcccccc}
\cmidrule[1pt](){1-7}
\multicolumn{1}{c}{\multirow{1}{*}{\hspace{0mm}\vspace{-7mm}\textbf{Methods / Metrics}}} & \multicolumn{3}{c}{\textbf{Accuracy} ($\uparrow$)} & \multicolumn{3}{c}{\textbf{Brier score} ($\downarrow$)}\\
\cmidrule(lr){2-4} \cmidrule(lr){5-7}
         &\textit{yes/no}&\textit{multi} &\textit{all} &\textit{yes/no} &\textit{multi} &\textit{all}     \\
\cmidrule[1pt]{1-7}
Random  &48.6    &25.3      &37.8      &0.684      &0.827      &0.750 \\
\midrule
Question      &66.2 &41.5      &54.7      &0.511      &0.715      &0.606 \\
Article   &73.6      &80.7      &76.9      &0.428      &0.263      &0.351 \\
Evidence sentence    &79.9    &89.5      &84.4      &0.355      &0.171      &0.269 \\
\cmidrule[1pt](){1-7}
\end{tabular}}
\caption{\small \textbf{Answerability study on test set.} 
Instead of retrieved articles, we provide \bertb with ground-truth context: a gold article or evidence sentence. We thus convert \data to a reading comprehension task and examine the answerability of the questions.
}
\label{result:gold}
\vspace{-0.1cm}
\end{table}

\para{Answerability of Questions.}
\label{sub-section:answerability}
To validate that the questions in \data are indeed answerable, we convert our setup into a machine reading comprehension (MRC) task --- find an answer given an assumed appropriate context.
We provide the model with a gold article or the evidence sentence
(Sec.~\ref{subsec:data_construction}).
Since pre-trained models have achieved high performance on MRC tasks \cite{Rajpurkar2016SQuAD10}, 
we expect adequate performance when provided the correct context.
As seen in Table~\ref{result:gold}, we observe that in closed-book setting, BERT is able to beat out a random baseline, but it still does not perform well; implying our questions are not trivial for BERT, and context is required to answer them correctly. 
When given the gold article, BERT achieves 76.9\% (+22\%) and it even performs better (84.4\%) given the evidence sentence.
This all implies that given the right information, our forecasting questions can be answered correctly.

\begin{figure}[tb!]
\vspace{-0.3cm}
    \centering
    \subfloat[Varying amounts of data.]{\includegraphics[width=0.45\columnwidth]{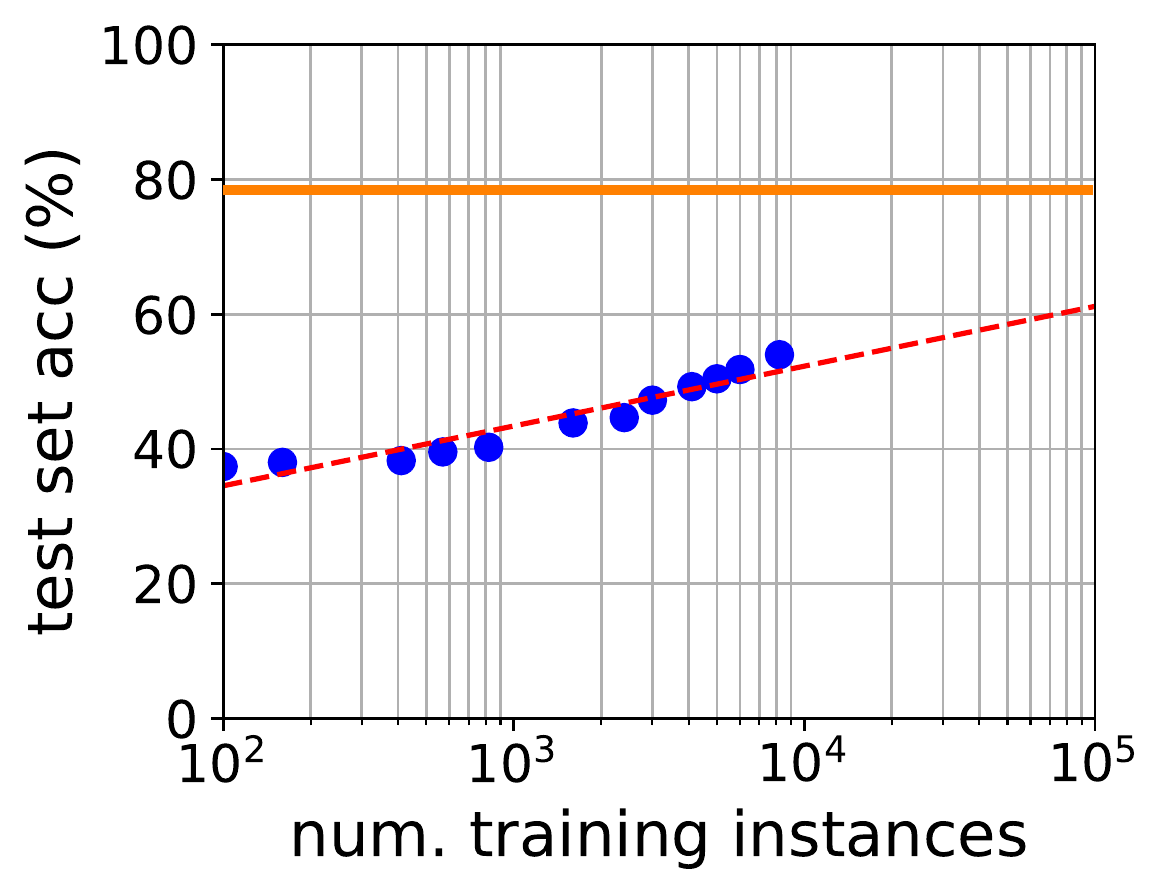} \label{fig:curve}}
    \quad
    \subfloat[Different question types.]{\includegraphics[width=0.45\columnwidth]{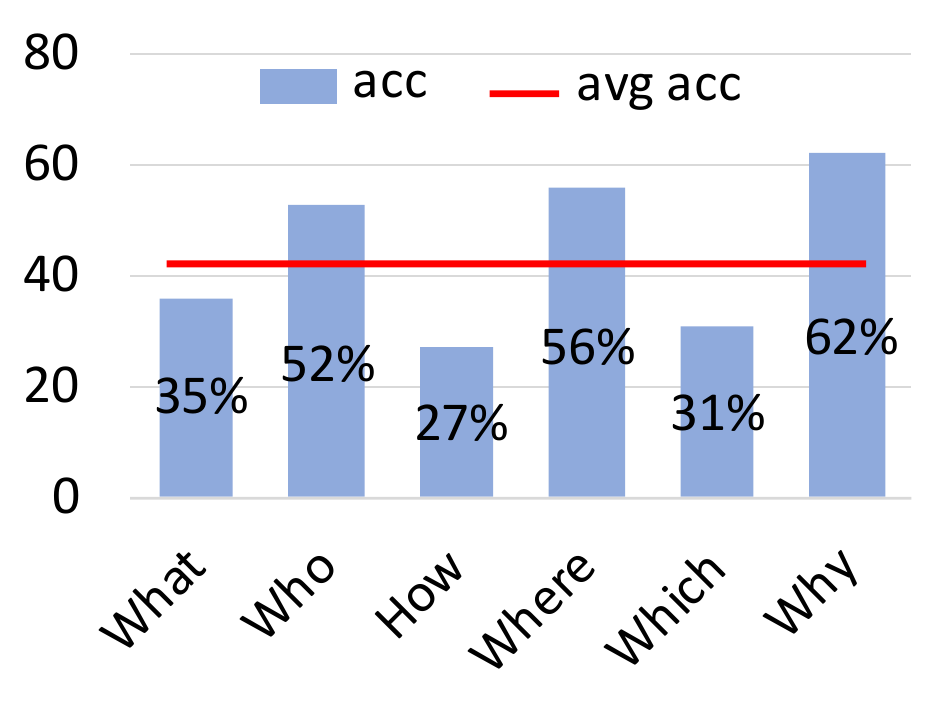} \label{fig:qtypeexp}}
    \caption{\small (a) Test accuracy of \bertb trained with varying amounts of training data, with human performance (79.1\%) shown in orange, and (b) development accuracy breakdown by different types of multichoice questions.}
    \label{fig:learningcurve}
    \vspace{0.0cm}
\end{figure}


\para{Study of Data Efficiency.}
To examine how models might perform with less/more training data, we evaluate \bertb (closed-book) on the test set, by training it with varying amounts of labeled data. Fig.~\ref{fig:curve} shows the the resulting ``learning curve." 
We observe the accuracy of the model is ``expected" to reach 70\%, assuming 100k examples --- which is still 9\%
point lower than human performance.

\para{Results on Different Question Types.}
We test \bertb (closed-book) on different question types of multi-choice questions from our development set (Fig.~\ref{fig:qtypeexp}). 
We find that the accuracy of the model varies  across different question types: ``\textit{how}" questions are the most difficult to predict while higher accuracy is achieved on ``\textit{why}" questions. 
Also for yes-no questions, the method achieves 69.5\% on ``\textit{yes}" questions and 62.9\% ``\textit{no}" questions, indicating that there is no significant bias towards certain type of binary questions.

\para{Error Analysis.} We observe 4 main categories of errors produced by the methods in our analysis: (1) retrieving irrelevant articles, (2) incorrect reasoning on relevant evidence, (3) lacking (temporal) common sense, and (4) lacking numerical knowledge. Please refer to Sec.~\ref{sec:error} of appendix for examples and in-depth discussions of these errors. 
\section{Conclusion}
Forecasting is a difficult task that requires every possible advantage to do well. 
It would be wise to harness this pool of unstructured data for training automatic event forecasting agents. To utilize this form of data for forecasting, we proposed a question-answering task that requires forecasting skills to solve \data, and provided the accompanying dataset. Various baseline methods did not perform well, but this is not surprising given the inherent difficulty of forecasting.
Our benchmark dataset can benefit future research beyond natural language understanding and hope forecasting performance will be significantly improved.

\bibliography{biblio}
\bibliographystyle{acl_natbib}

\clearpage
\appendix

\section{Detailed Dataset Creation}
\label{append:data}
In this section, we present detailed explanations of dataset creation.
We first selected news sources as in the following section.

\subsection{List of News Sources}
The New York Post, The New York Times, New York Magazine, Daily News (New York), The Washington Post, NPR All Things Considered, NPR Weekend Edition Saturday, NPR Morning Edition, CNN Wire, CNN.com, CNNMoney.com, CNN INTERNATIONAL, Fox News Network, York Guardian, Washingtonpost.com, The Washington Post Magazine, thetimes.co.uk, Guardian Weekly, Russia \& CIS General Newswire, US Official News, The Times (London).

\begin{figure}[tb!]
\centering
\includegraphics[width=1\linewidth]{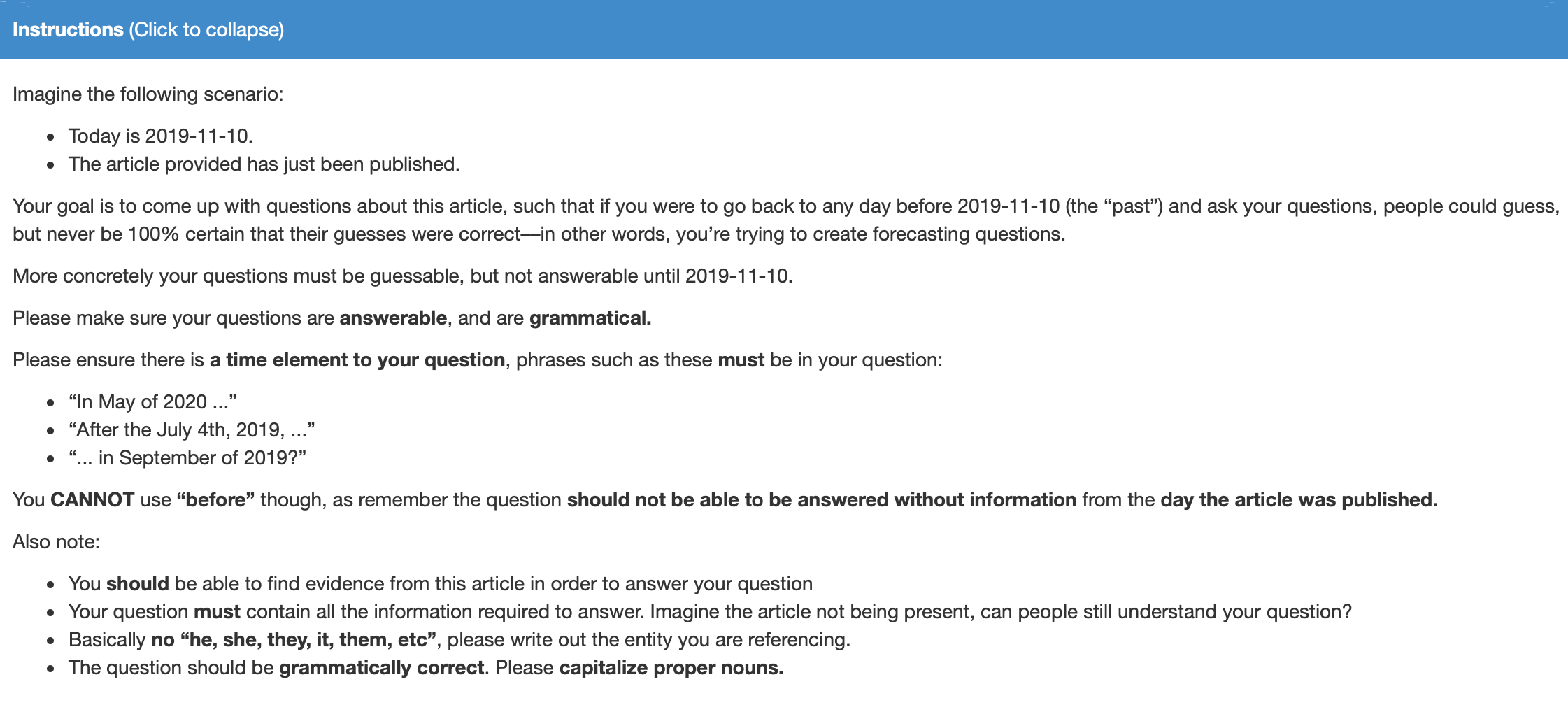}
\caption{Instruction of creating multiple-choice questions.}
\label{fig:int:instruction}
\end{figure}

\begin{figure}[tb!]
\centering
\includegraphics[width=1\linewidth]{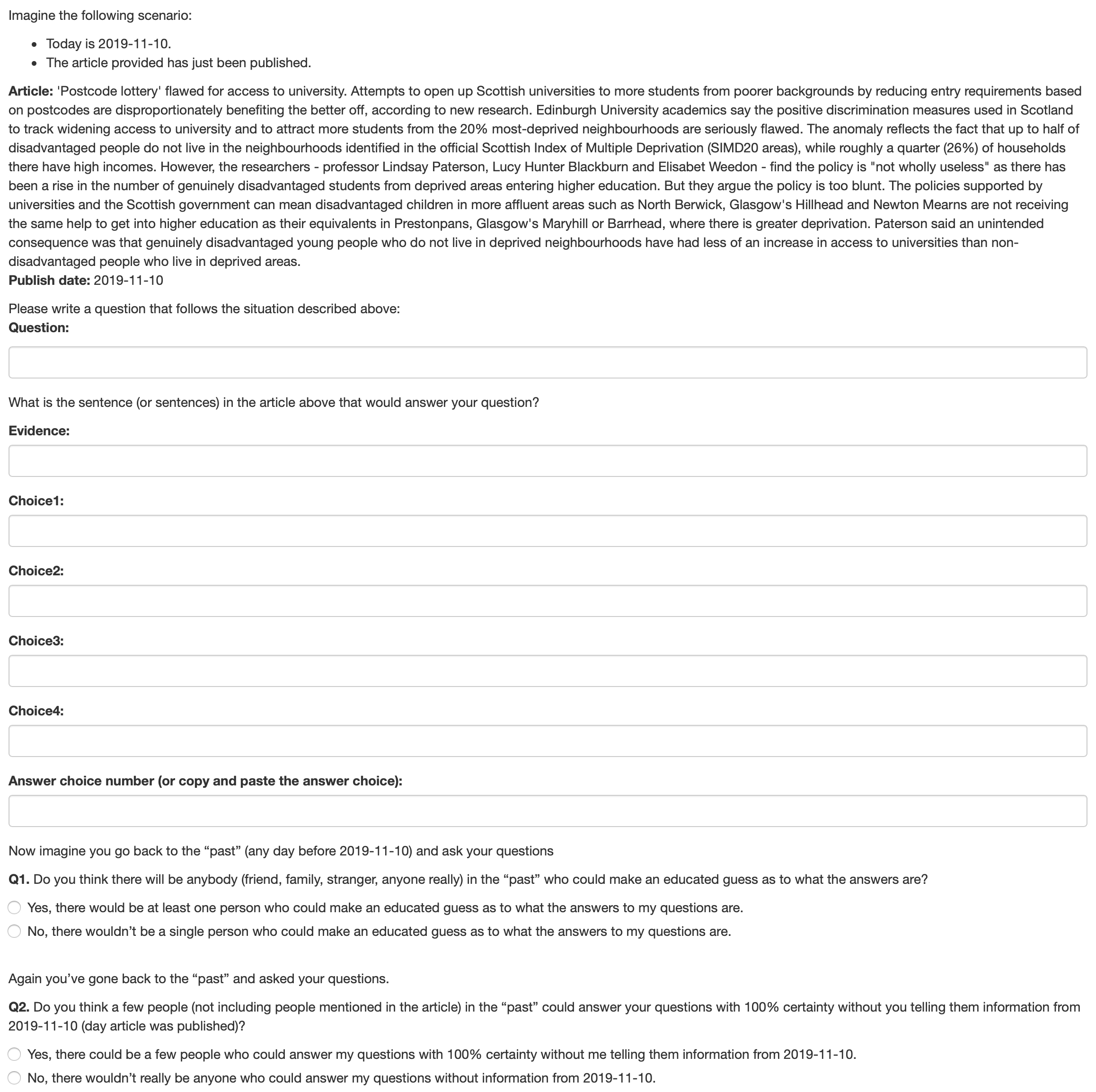}
\caption{Interface of creating multiple-choice questions.}
\label{fig:int:creation}
\end{figure}

\begin{figure}[tb!]
\centering
\includegraphics[width=1\linewidth]{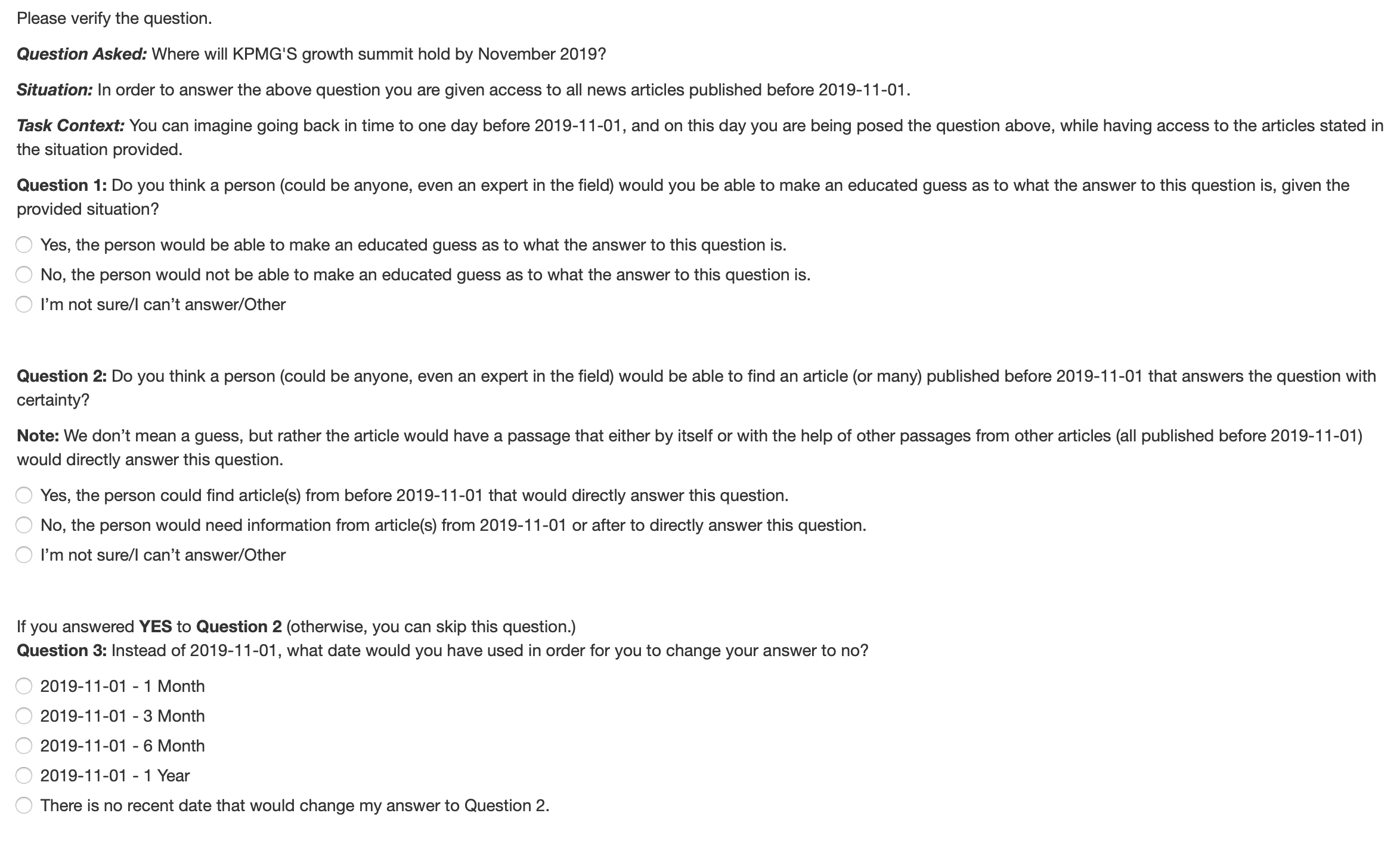}
\caption{Interface of verifying questions.}
\label{fig:int:verify}
\end{figure}

\subsection{Dataset Creation}
\para{Turking Guidelines.}
Figs~\ref{fig:int:instruction} and \ref{fig:int:creation} show the instructions and interface for creating our multiple-choice questions.
Workers made multiple-choice distractors with their own knowledge, but they were encouraged to find good distractors using search engines. 
To ensure the answerability of the created questions, we ask them to indicate the answer along with the supporting evidence that the question is made from.
We omit the interfaces due to the space limit.

\para{Initial Screening.}
The ideal result of our crowdsourcing task are forecasting questions that are tractable but not trivial, and by definition not answerable with certitude using information currently available. Thus to avoid undesirable questions, we asked two additional questions to help screen poorly constructed questions.
As shown in Fig~\ref{fig:int:creation}, we try to determine the difficulty of the question and whether it is answerable using ``current" or ``past" information. Question 1 attempts to establish whether the question is indeed tractable and asks whether there exists some qualified group of people who could reason and make an educated guess at the answer to the question. On the other hand, question 2 tries to determine if the question is either too easy or is definitively answerable given ``current" and ``past" information. Thus, the desired response is ``yes" and ``no" for Questions 1 and 2, respectively; we filtered out created questions that do not satisfy the desired condition.



\subsection{Additional Question Quality Checks}
We asked the same two questions from our initial quality screening and an additional question to help adjust the timestamp associated with the question if needed. Per question, we got 3 crowd workers to answer the three questions and took the majority vote for question 1 and 2, while selecting the earliest selected timestamp for question 3. We dropped the question, if the majority vote was ``no" for question 1  or  ``yes'' for question 2.
Moreover, if at least one worker selected ``e" in the question 3 (There is no appropriate recent time stamp), then we filtered out the question.
Additionally, if the created question does not have a temporal phrase, then we filter out the question.


\begin{table}[tb!]
\centering
\resizebox{1\linewidth}{!}{
\begin{tabular}{p{14cm}}
\toprule
\textbf{Q:} What wild animal will be found at the Outer banks of North Carolina in September 2019? \\
\textbf{Choices:} \textbf{Horses} (\textit{answer}), Cows, Turtles, Donkeys. \\
\cmidrule[1pt]{1-1}
\textbf{Article:}  \textit{Tillis Introduces Legislation to Protect Corolla Wild Horses Washington: Office of the Senator Thom Tillis has issued the following news release:} (1/29/19) \newline U.S. Senator Thom Tillis (R-NC) introduced the Corolla Wild Horses Protection Act, legislation that would provide responsible management of the wild horse population around Corolla, North Carolina and the Outer Banks. Representative Walter Jones (R-NC) introduced companion legislation in the House of Representatives in previous Congresses and has been a long time champion of protecting the Corolla wild horse population. \\
\cmidrule[0.5pt]{1-1}
\textbf{Reasoning Process:} The Corolla Wild Horses Protection Act will make people to protect the wild horses (\textbf{forecasting skills - causal relations}). If people start to protect the wild horses from January, the wild horses will be found in September (\textbf{forecasting skills - inferring based on past events} - \textit{we can find the answer from this part}). Horse is an animal (\textbf{commonsense - world knowledge}). The Outer banks of North Carolina = North Carolina and the Outer Banks (\textbf{language understanding - paraphrase}). \\
\bottomrule
\end{tabular}}
\caption{Detailed example to show how to solve a question. 
}
\label{table:more_detailed_example2}
\end{table}

\section{Example of Reasoning}

Table~\ref{table:more_detailed_example2} shows an example of reasoning process to solve a question.

\begin{figure}[tb!]
    \centering
    \includegraphics[width=0.9\linewidth]{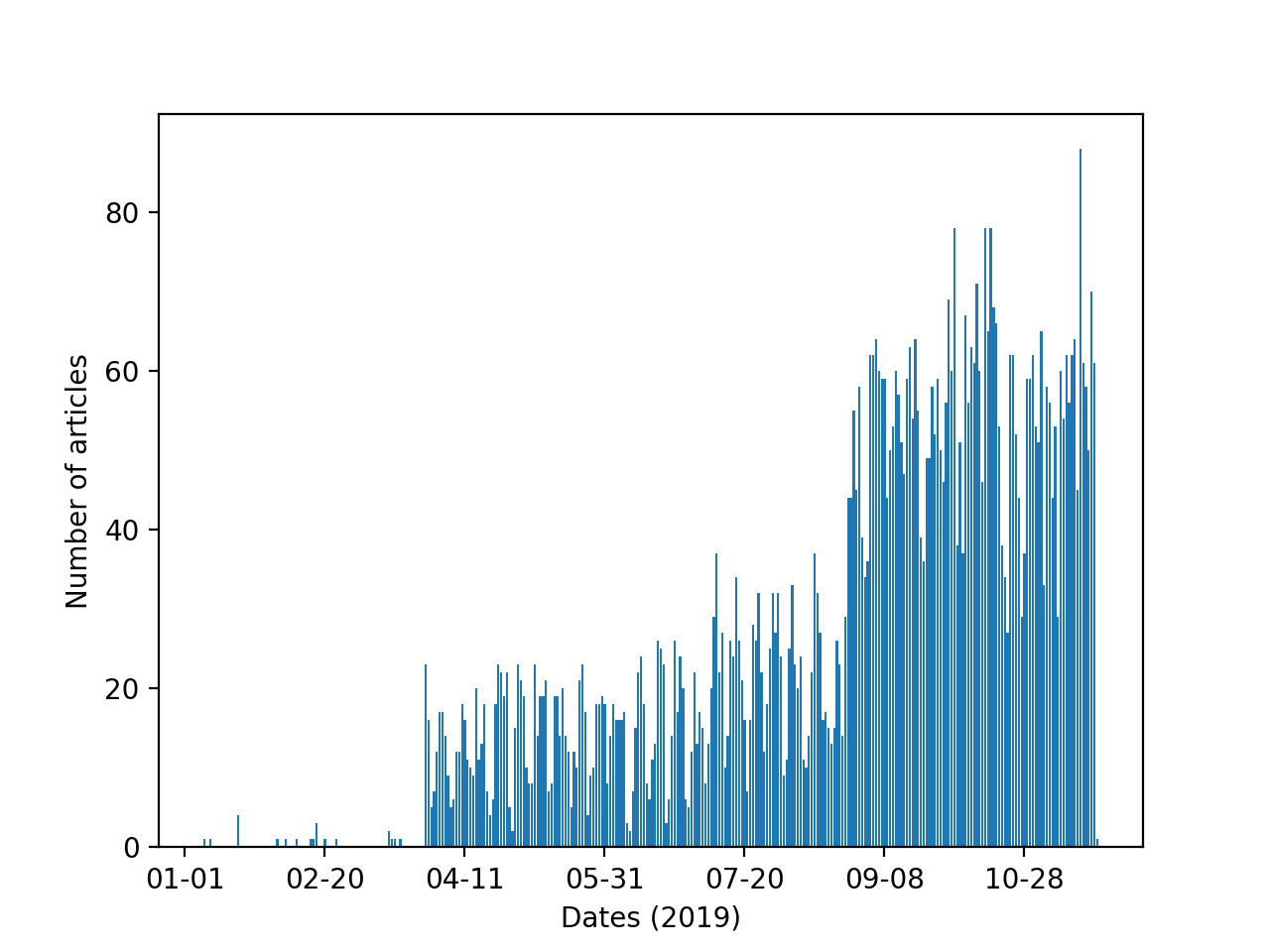}
    \caption{Date distribution of gold articles for questions. Each question is made from gold articles. The dates denote release dates of news articles and they range from 01-01-2019 to 11-31-2019. 
    }
    \label{fig:dist}
\end{figure}

\begin{figure}[tb!]
    \centering
    \includegraphics[width=1\linewidth]{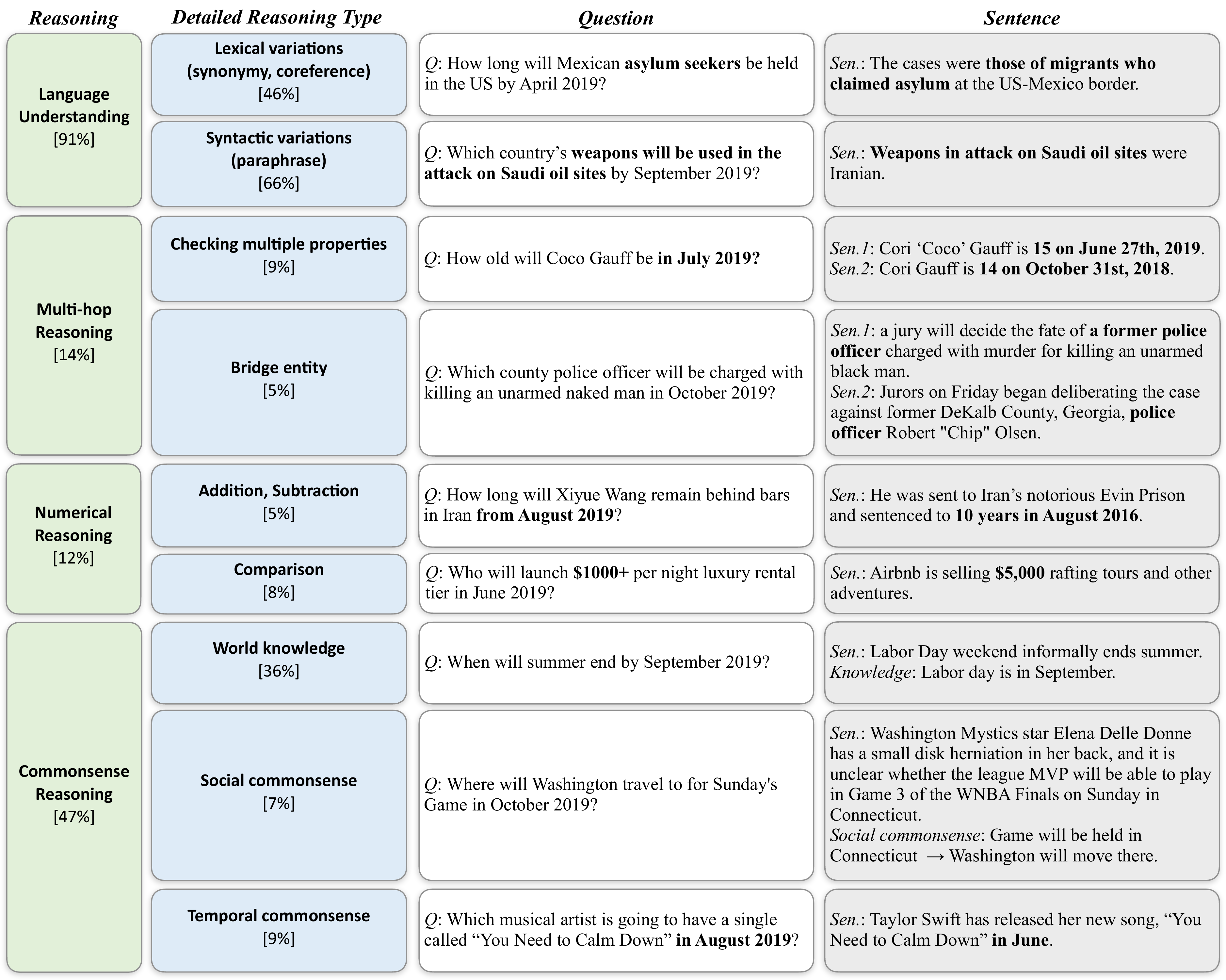}
    \caption{Examples of each type of reasoning in \data. Words relevant to the corresponding reasoning type are bolded. Also, [\%] represents the percentage of questions that requires the reasoning type. 
    }
    \label{fig:reasoning_example2}
\end{figure}

\section{Additional Reasoning Types}
\label{appendix:reasoningtypes}

Figure~\ref{fig:reasoning_example2} shows additional reasoning types.

\para{Language Understanding.} 
We introduce lexical variations and syntactic variations following \citet{Rajpurkar2016SQuAD10,Rajpurkar2018KnowWY}.
Lexical variations represent synonyms or coreferences between the question and the evidence sentence.
When the question is paraphrased into another syntactic form and the evidence sentence is matched to the form, we call it syntactic variation.
We find that many questions require language understanding; lexical variations account for 46\% and syntactic variations do for 66\%. 

\para{Multi-hop Reasoning.} 
Some questions require multi-hop reasoning~\cite{Yang2018HotpotQAAD}, such as checking multiple properties (9\%) and bridge entities (5\%) .
The former one requires finding multiple properties from an article to find an answer.
The latter one works as a bridge between two entities, where one must identify a bridge entity, and find the answer in the second hop.

\para{Numerical Reasoning.} 
To answer our questions, one needs numerical reasoning~\cite{Dua2019DROPAR}.
The answer is found by adding or subtracting two numbers (5\%), or comparing two numbers (8\%) in the given articles.

\para{Commonsense Reasoning.} 
The questions require world knowledge~\cite{Talmor2018CommonsenseQAAQ}, social commonsense~\cite{Sap2019SocialIC}, and temporal commonsense~\cite{Zhou2019GoingOA}.
To solve these questions, an AI agent must leverage assumed common knowledge in addition to what it finds in the news corpus. We find that 36\% questions need world knowledge and 7\% questions require social commonsense.
The other type of commonsense reasoning is temporal commonsense, which is related to temporal knowledge~\cite{Zhou2019GoingOA}.
9\% questions are related to temporal commonsense.

\begin{table}[tb!]
\centering
\resizebox{0.95\columnwidth}{!}{
\begin{tabular}[t]{L{8cm}|C{2cm}}
\toprule
\textbf{Measurement}    &\textbf{Value}\\
\cmidrule[1pt]{1-2}
Average question length (tokens)    & 13.85  \\
Average answer length (tokens)  & 2.46  \\
\# of distinct words in questions    &  17,521 \\
\# of distinct words in choices  &   5,187\\
\# of distinct time stamps associated w. questions    &   218 \\
Average gold article length (\# tokens)  & 720.21  \\
Maximum question time stamp  &   2019-11-22\\
Minimum question time stamp  &   2019-01-01\\
\bottomrule
\end{tabular}}
\caption{Statistics of \data. 
}
\label{data:keystats}
\end{table}

\begin{table}[tb!]
\centering
\resizebox{1\columnwidth}{!}{
\begin{tabular}[t]{L{2.5cm}R{1cm}L{5cm}}
\toprule
\textbf{Answer Type}    &\textbf{\%}  & \textbf{Examples} \\
\cmidrule[1pt]{1-3}
Yes/No  & 56.8\%  & -\\
Person  &   8.1\%& Boris Johnson, Mark Zuckerberg\\
Group/Org  &   5.8\%& BBC, United Nations, EU\\
Location  &   8.0\%& Canada, Iran, U.S.\\
Date/Time  &   1.6\%& January, July\\
Number  &   6.7\%& 530, Thirty eight\\
Other Entity    &   1.1\%&  Boeing 737 \\
Common Noun Phrase    &   5.8\%& A hurricane, Asteroid dust   \\
Verb Phrase    &   3.1\%&  Defend his innocence \\
Adjective Phrase    &   1.4\%&   Cruel and Misguided, Due to the bad weather\\
Sentence    &   1.6\%&Liverpool will become the first English team to play their 400th international game.\\
\bottomrule
\end{tabular}}
\caption{Types of answers in \data. 
}
\label{data:answertype}
\end{table}

\section{Statistics}
\label{append:stat}
Tables \ref{data:keystats} and \ref{data:answertype} show the statistics and answer types in \data.


\section{Experiments}
\subsection{Details on a Text Encoder}
We use Huggingface's codes\footnote{\url{https://github.com/huggingface/transformers}}.
We chose the best learning rate among $\{3e-5, 1e-5, 5e-6\}$ and the number of epochs is 3. 
We set the max sequence length to 512.

\subsection{Details on IR methods}
\label{exp:ir}
We index the English news articles with Elasticsearch~\cite{gormley2015elasticsearch}.
We followed the setups in \citet{Qi2019AnsweringCO}.
We use Elasticsearch’s simple analyzer which performs basic tokenization and lowercasing for the title.
We use the standard analyzer which allows for removal of punctuation and stop words from the body of articles.
At retrieval time, 
we use a \texttt{multi\_match} query in the Elasticsearch against all fields with the same query, which performs a full-text query employing the BM25 ranking function~\cite{robertson1995okapi} on all fields, and returns the score of the best field for ranking. 
To promote documents whose title matches the search query, we boost the search score of any result whose title matches the search query by 1.25, which results in a better recall for entities with common names.

\subsection{Details on Baselines.}
\label{exp:baselines}
We consider following baselines:
(1) \textbf{Event-based approaches}: We test event-based approach, BERT with event triples (two entities and a relation between them) and BERT based on SAM-Net~\cite{lv2019sam} for our setup. It is non-trivial to apply the event-based approaches to our setup. Thus, we preprocess the retrieved news articles into event triples (subject, relation, object) using \citet{liu2019open}. We simply regard them as text, we concatenate the triples, and feed them into BERT and call it \textbf{BERT with event triples}.
In addition, we apply a script learning approach (SAM-Net~\cite{lv2019sam}) to our setup. 
A question and choices are not used in their original method; thus we encode them using BERT and concatenate the encodings with the approach's final representation. 
This representation is fed into a linear layer and the linear layer predicts whether the choice is correct or not. 
We used \bertl for the former one and \bertb for the latter one.
(2) \textbf{ESIM} \cite{Chen2017EnhancedLF}. An NLI model, where we change their output layer so that the model outputs probabilities for each answer choice with a softmax layer. 
We use ELMo~\cite{Peters2018DeepCW} for word embeddings.
(3) \textbf{\textsc{BiDAF++}}~\cite{Clark2018SimpleAE}. The model requires context, and thus we use a top-1 article by an IR method.
We augment it with a self-attention layer and ELMo representations.
To adapt to the multiple-choice setting, we choose the answer with the highest probability.
The input to ESIM is a question and a set of choices $(Q,C)$, while that of \textsc{BiDAF++}'s is a question, a set of choices, and retrieved articles $(Q,C,\overline{A})$.\footnote{We did not include existing event forecasting methods since they are designed for modeling structured event data~\cite{fawaz2019deep} and thus are not directly applicable to \data which requires modeling of unstructured text.}

\subsection{Time Modeling}
\label{exp:time}
We conduct an ablation study on modeling time information of the retrieved articles.
We test the following models:
a) pre-pending date string as BERT input $[\texttt{[CLS]} Q \allowbreak \texttt{[SEP]} C \texttt{[SEP]} \textit{Date} \texttt{[SEP]} A_i]$, where the date format is ``YYYY-MM-DD",
b) using binary encodings of dates: we first encode the time into a binary encoding using ``Temporenc\footnote{\url{https://temporenc.org}}" and concatenate the encoding with an article encoding before aggregation,
c) using char-RNN~\cite{goyal2019embedding} for encoding date string before aggregation.

\begin{table}[tb!]
\centering
\resizebox{0.9\linewidth}{!}{
\begin{tabular}{lC{1.5cm}C{1.5cm}C{1.5cm}}
\cmidrule[1pt](){1-4}
\multicolumn{1}{c}{\multirow{1}{*}{\hspace{0mm}\vspace{-7mm}\textbf{Methods / Metrics}}} & \multicolumn{3}{c}{\textbf{Accuracy}} \\
\cmidrule(lr){2-4} 
         &\textit{yes/no}&\textit{multi} &\textit{all}    \\
\cmidrule[1pt]{1-4}
\bertb, AGG (GRU)  &67.6    &41.5      &55.4\\
RoBERTa\textsubscript{BASE}, AGG (GRU)  &69.3   	&44.8 	&57.9\\
AL\bertb, AGG (GRU)	  &67.4   	&23.4 	&46.9\\
\midrule
\bertl, AGG (GRU)  &69.2    &47.5      &59.1\\
RoBERTa\textsubscript{LARGE}, AGG (GRU)  &70.1   	&51.3 	&61.3\\
AL\bertl, AGG (GRU)	  &68.4   	&30.2 	&50.6\\
\midrule
Human performance		  &81.3		&77.4 	&79.4\\
\cmidrule[1pt](){1-4}
\end{tabular}}
\caption{Results on different pre-trained language models, BERT, RoBERTa, ALBERT). 
}
\label{table:lms}
\end{table}

\subsection{Experiments with Recent LMs.}
\label{append:stronglms}
As mentioned in Sec~\ref{sec:models}, we did not report more recent pre-trained language models (e.g., RoBERTa~\cite{Liu2019RoBERTaAR}, ALBERT~\cite{lan2019albert}) because they are trained using text data published after the earliest timestamp in our dataset (2019-01-01). We are worried that these models in theory would have access to information that was published after the associated timestamp of a question.

As a reference, we show the results of RoBERTa and ALBERT in Table~\ref{table:lms}. Even though these two models may violate our forecasting scenario, they still struggle when compared to human performance, suggesting that our task is still challenging.

\begin{table}[tb!]
\centering 
\resizebox{1.0\linewidth}{!}{
\begin{tabular}{lcccccc}
\cmidrule[1pt](){1-7}
\multicolumn{1}{c}{\multirow{1}{*}{\hspace{0mm}\vspace{-7mm}\textbf{Methods / Metrics}}} & \multicolumn{3}{c}{\textbf{Accuracy (\%,~$\uparrow$)}} & \multicolumn{3}{c}{\textbf{Brier score ($\downarrow$)}}\\
\cmidrule(lr){2-4} \cmidrule(lr){5-7}
         &\textit{yes/no}&\textit{multi} &\textit{all} &\textit{yes/no} &\textit{multi} &\textit{all}     \\
\cmidrule[1pt]{1-7}
\bertl, AGG (GRU)  &69.2   &47.5   &59.1   &0.483   &0.655    &0.563    \\
\bertl, GRU(A), QC  &67.8    &42.5   &56.0   &0.583   &0.758   &0.665\\
\cmidrule[1pt](){1-7}
\end{tabular}}
\vspace{-0.1cm}
\caption{ Performance of baseline models on \data test set.
}
\label{result:open2}
\vspace{-0.1cm}
\end{table}

\subsection{Experiments with different GRU architectures.}
We investigate GRU modeling for the input. \bertl GRU(A), QC refers to a model that encodes each article with a text encoder, these encodings are fed into GRU, and concatenate the last hidden representation of GRU and Q,C (question and choice) encoding from the text encoder.
Table~\ref{result:open2} shows comparison between the two architectures. Separating the articles with the question and choice leads to the worse performance.

\begin{table}[tb!]
\centering
\resizebox{0.95\linewidth}{!}{
\begin{tabular}{lcccccc}
\cmidrule[1pt](){1-7}
\multicolumn{1}{c}{\multirow{1}{*}{\hspace{0mm}\vspace{-7mm}\textbf{Methods / Metrics}}} & \multicolumn{3}{c}{\textbf{Accuracy (\%)}} & \multicolumn{3}{c}{\textbf{Brier score}}\\
\cmidrule(lr){2-4} \cmidrule(lr){5-7}
         &\textit{yes/no}&\textit{multi} &\textit{all} &\textit{yes/no} &\textit{multi} &\textit{all}     \\
\cmidrule[1pt]{1-7}
\textbf{\bertb}  &&&&&&\\
~~$-$~Question      &65.6 &43.7      &55.4      &0.506      &0.698      &0.596 \\
~~$-$~Article  &78.1  &84.8      &81.2      &0.351      &0.210      &0.285\\
~~$-$~Evidence sentence    &81.4    &90.5      &85.6      &0.324      &0.147      &0.241 \\
\cmidrule{1-7}
\cmidrule[1pt](){1-7}
\end{tabular}}
\caption{Results on gold articles on the dev set. 
We give different inputs to the BERT to find out which part is important for the questions. 
}
\label{result:gold2}
\end{table}

\subsection{Error Analysis}
\label{sec:error}
\begin{figure}[tb!]
\vspace{-0.2cm}
    \centering
    \includegraphics[width=0.9\linewidth]{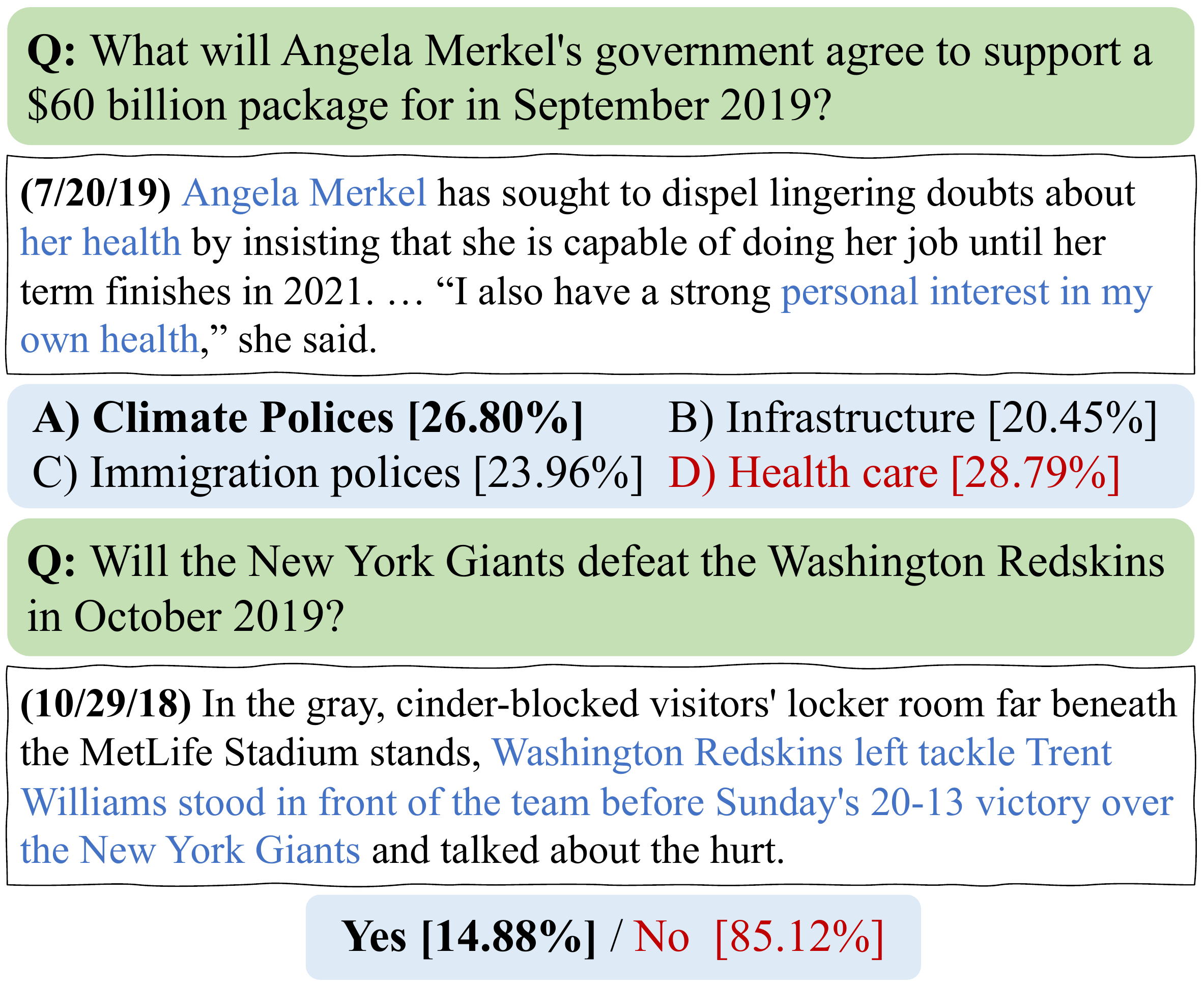}
    \caption{Examples of erroneous model predictions. Bold choices are actual answers and red choices are model predictions. 
    }
    \label{fig:error_anal}
    \vspace{0.2cm}
\end{figure}

We randomly select 50 errors made by the best baseline method from the test set and identify 4 phenomena:

\para{Retrieving Wrong Articles.}
28\% of the errors are from the retrieval of irrelevant articles.
The baseline approach relies on information retrieval methods such as BM25. 
Retrieved articles might not be relevant or contain facts that can confuse the model, thus causing incorrect predictions. For example, consider the first question in Fig.~\ref{fig:error_anal}, the model has retrieved an irrelevant article and conflated Ms. Merkel's health with policy decisions. This results in the model incorrectly choosing Health Care as the appropriate answer.

\para{Incorrect Use of Relevant Evidence.}
24\% of the errors are (partially) caused by incorrect usage of relevant evidence.
Even though useful articles are retrieved, the model incorrectly reasons over the evidence.
Take the second question in Fig.~\ref{fig:error_anal}, where the model incorrectly predicts \textit{No}. The model may depend on a relevant, but outdated fact from 2018 (one year before the event in question) to answer the question, and failed to incorporate more recent information.

\para{Lacking Human Common Sense.}
32\% of the errors are from the model's lack of common sense or world knowledge.
An example question is, \textit{``Who will host 2020 Olympics by July 2019?,"} where the answer is Japan, but the model predicts Hong Kong. To answer this question, a model must know the cities of each country, as without this knowledge the model does not know that ``Tokyo is in Japan," and thus the model predicts the wrong answer.

\para{Numerical Questions.}
8\% of the errors are from numerical questions.
Numerical questions ask about numbers such as a person's age. 
For example, \textit{``What will be Roger Federer's age by August 2019."} The model must know his birth month and age and know how to increment on one's birthday.

\end{document}